\documentclass[conference]{IEEEtran}
\IEEEoverridecommandlockouts
\usepackage{cite}
\usepackage{dblfloatfix}
\usepackage{amsmath,amssymb,amsfonts}
\usepackage[absolute]{textpos}
\usepackage{graphicx}
\usepackage{textcomp}
\usepackage{xcolor}
\usepackage{algorithm}
\usepackage{algpseudocode}
\usepackage{arabtex}
\usepackage{utf8}
\usepackage[utf8]{inputenc}
\setcode{utf8}
\usepackage{url}
\usepackage{makecell}
\usepackage{hyperref}
\usepackage{subcaption}
\usepackage{scalerel,xparse}
\usepackage{multirow}
\usepackage{fancyhdr}
\usepackage{multirow}%
\usepackage{tabularx}
\usepackage{booktabs}%
\hypersetup{%
	pdfborder = {0 0 0}
}
\def\BibTeX{{\rm B\kern-.05em{\sc i\kern-.025em b}\kern-.08em
    T\kern-.1667em\lower.7ex\hbox{E}\kern-.125emX}}
\begin{document}
\makeatletter
\def\ps@IEEEtitlepagestyle{%
	\def\@oddfoot{\mycopyrightnotice}%
	\def\@evenfoot{}%
}
\def\mycopyrightnotice{
	{\footnotesize 979-8-3503-5026-5/24/\$31.00~\copyright2024 IEEE \hfill}
	\gdef\mycopyrightnotice{}
}
\title{Uterine Ultrasound Image Captioning Using Deep Learning Techniques}
\NewDocumentCommand\emojismiley{}{
	\includegraphics[scale=0.2]{1f914.png}
}
\newcommand{\linebreakand}{%
\end{@IEEEauthorhalign}
\hfill\mbox{}\par
\mbox{}\hfill\begin{@IEEEauthorhalign}
}
\author{
\IEEEauthorblockN{Abdennour Boulesnane$^*$\href{https://orcid.org/0000-0002-2272-4953}{\includegraphics[scale=0.02]{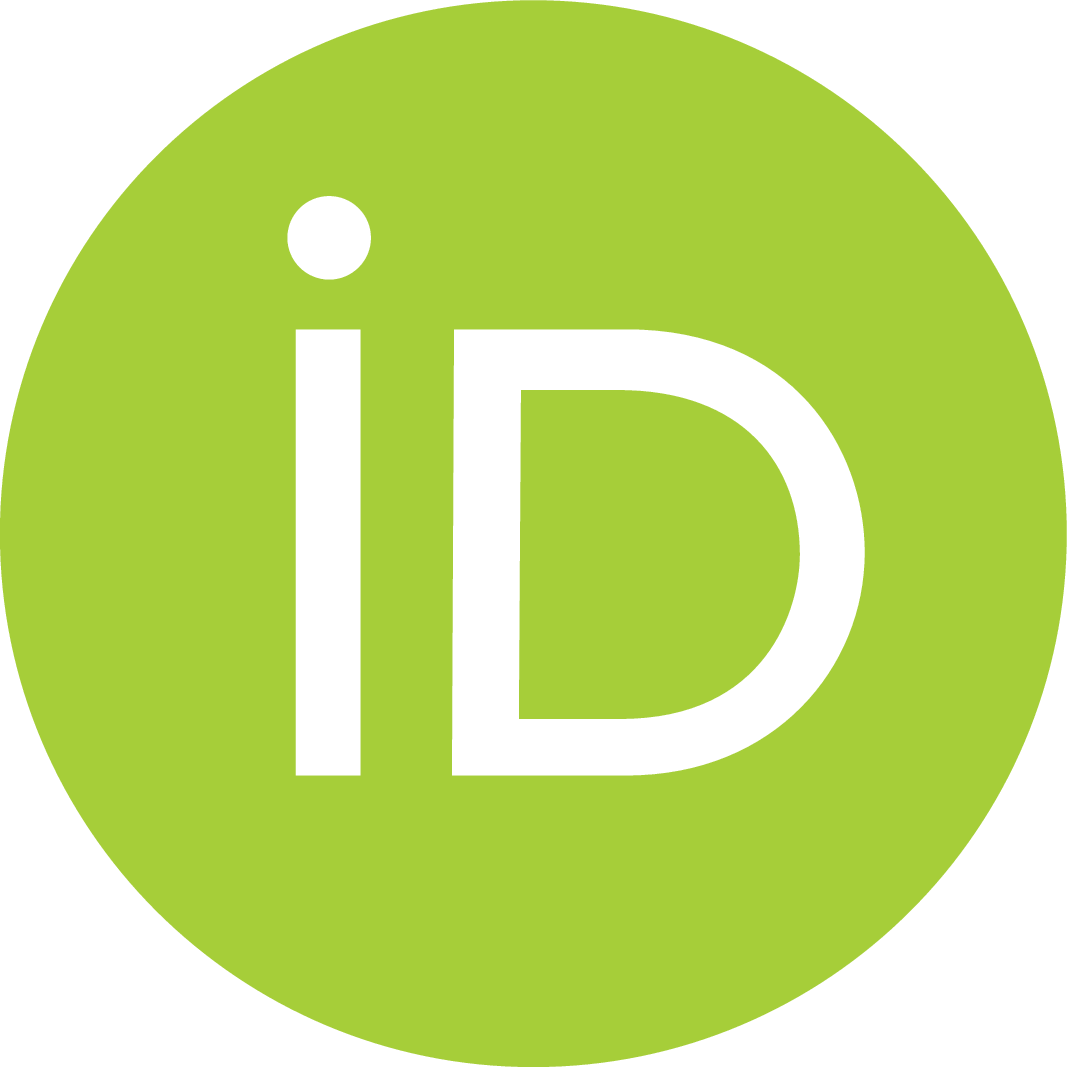}}}
\IEEEauthorblockA{BIOSTIM Laboratory \\
	Faculty of Medicine \\
	Salah Boubnider University\\
	Constantine, Algeria \\
	aboulesnane@univ-constantine3.dz}
\and		
\IEEEauthorblockN{Boutheina Mokhtari$^*$}
\IEEEauthorblockA{Department of IFA\\
	Faculty of NTIC \\
	Abdelhamid Mehri University\\
	Constantine, Algeria \\
boutheina.mokhtari@univ-constantine2.dz}
\and	
\IEEEauthorblockN{Oumnia Rana Segueni$^*$}
\IEEEauthorblockA{Department of IFA\\
	Faculty of NTIC \\
	Abdelhamid Mehri University\\
	Constantine, Algeria \\
	oumnia.segueni@univ-constantine2.dz}
\linebreakand 
\IEEEauthorblockN{Slimane Segueni}
\IEEEauthorblockA{Obstetrics and Gynecology Clinic\\
23 Khelifi Abderrahmane Street\\Chelghoum Laid, Mila, Algeria \\
segueni\_slimane@yahoo.fr}	
}
\maketitle
\def\thefootnote{*}\footnotetext{The authors contributed equally to this work.}
\thispagestyle{fancy}
\cfoot{}
\rfoot{}
\renewcommand{\headrulewidth}{0pt}
\renewcommand{\footrulewidth}{0pt}
\pagestyle{fancy}
\cfoot{\thepage}
\begin{abstract}
Medical imaging has significantly revolutionized medical diagnostics and treatment planning, progressing from early X-ray usage to sophisticated methods like MRIs, CT scans, and ultrasounds. This paper investigates the use of deep learning for medical image captioning, with a particular focus on uterine ultrasound images. These images are vital in obstetrics and gynecology for diagnosing and monitoring various conditions across different age groups. However, their interpretation is often challenging due to their complexity and variability. To address this, a deep learning-based medical image captioning system was developed, integrating Convolutional Neural Networks with a Bidirectional Gated Recurrent Unit network. This hybrid model processes both image and text features to generate descriptive captions for uterine ultrasound images. Our experimental results demonstrate the effectiveness of this approach over baseline methods, with the proposed model achieving superior performance in generating accurate and informative captions, as indicated by higher BLEU and ROUGE scores. By enhancing the interpretation of uterine ultrasound images, our research aims to assist medical professionals in making timely and accurate diagnoses, ultimately contributing to improved patient care.
\end{abstract}


\section{Introduction}\label{sec1}
Throughout the long and rich history of medical imaging, significant advancements have revolutionized diagnostic and treatment methodologies \cite{Haidekker2013}. Beginning over a century ago with the invention of X-rays, which allowed non-invasive visualization of the human body, the field has continually evolved. Modern technologies such as MRIs, CT scans, Positron Emission Tomography (PET), and ultrasounds enable precise diagnosis and treatment across various medical conditions, enhancing our understanding of human anatomy \cite{bradley2008history}. Recently, the integration of computer vision, natural language processing (NLP), and artificial intelligence (AI) has further transformed medical imaging, paving the way for unprecedented developments \cite{Obuchowicz2024}.\\
Deep learning-based AI technologies, in particular, have demonstrated remarkable potential in diagnosing patients swiftly and accurately \cite{Suzuki2017}. These automated machine learning algorithms significantly alleviate the workload of medical professionals, offering the promise of transformative impacts on healthcare and patient care \cite{Kaul2021}. One of the most captivating advancements in this domain is medical image captioning (MIC) \cite{Beddiar2022}. By leveraging deep learning, MIC systems can automatically generate captions for medical images, combining expert annotations with images from extensive datasets to provide precise and detailed analyses. These capabilities enhance medical documentation, expedite diagnosis, and facilitate remote consultations, thereby improving overall healthcare delivery \cite{Xu2023}.

Despite these advancements, interpreting medical images remains a formidable challenge \cite{Xu2023}. Variability in doctors' experience levels can lead to inconsistent diagnoses, and misunderstandings can result in medical errors that adversely affect patient outcomes. Moreover, reading and analyzing these images can be time-consuming, particularly in emergencies where rapid decision-making is crucial. Uterine ultrasound images, in particular, present unique challenges in obstetrics and gynecology. Their generally lower quality than other medical images complicates interpretation, potentially leading to delayed or incorrect diagnoses and impacting patient care \cite{LevienaiseObadia1999}. The complexity and variability of these images underscore the need for an effective MIC system, serving as the primary motivation for our research.

Our study aims to address the challenges in interpreting uterine ultrasound images by developing a specialized MIC system to enhance diagnostic precision and efficiency. We assembled a comprehensive dataset of uterine ultrasound images, prioritizing patient privacy and confidentiality. This dataset was meticulously annotated with expert-provided descriptions, ensuring high-quality data for training and evaluation. We then undertook extensive data preprocessing, isolating regions of interest within the images using a cropping algorithm and standardizing the text annotations with NLP techniques. This preparation laid a solid foundation for feature extraction and model development.\\
In the feature extraction stage, we employed pre-trained Convolutional Neural Network (CNN) models such as Inception V3 and DenseNet201 to obtain detailed feature vectors from the images. Concurrently, we transformed the textual data into numerical representations to enable seamless integration with the image features. Our deep learning model combines these processed inputs through a Bidirectional Gated Recurrent Unit (BiGRU) network, generating descriptive captions for the ultrasound images. Evaluated using metrics like BLEU and ROUGE scores, our CNN-BiGRU model demonstrated promising results in accurately describing uterine ultrasound images. These findings underscore the effectiveness of our approach and its potential to enhance diagnostic accuracy in gynecology, ultimately contributing to improved patient care.

The remainder of the paper is structured as follows: Section \ref{sec2} presents a review of related works. Section \ref{sec3}  details the proposed approaches. In Section \ref{sec4}, we analyze and discuss the experimental results. Finally, Section \ref{sec5} offers conclusions and outlines directions for future research.
\section{Related Work}\label{sec2}
Ultrasound imaging is invaluable for visualizing complex anatomical structures, offering advantages such as portability, real-time imaging, cost-effectiveness, and the absence of radiation \cite{Chen2016}. However, interpreting these images can be challenging due to their often low quality, with common issues such as fuzzy borders and numerous artifacts \cite{Zeng2018}. While numerous studies have focused on medical image captioning (MIC) \cite{Beddiar2022}, the majority target medical reports for chest X-ray images \cite{Wang2024}, leaving MIC for ultrasound images relatively underexplored. This section will delve into MIC research specifically pertaining to ultrasound images.

In \cite{Zeng2018}, a coarse-to-fine ensemble model for ultrasound image captioning is presented. The model first detects organs using a coarse classification model, then encodes the images with a fine-grained classification model, and finally generates annotation text describing disease information using a language generation model. The model, trained using transfer learning from a pre-trained VGG16 model, achieves high accuracy in ultrasound image recognition.\\
Building on the concept of combining different models, \cite{Alsharid2019} introduces an NLP-based method to caption fetal ultrasound videos using vocabulary typical of sonographers. This approach combines a CNN (based on VGGNet16, fine-tuned on fetal ultrasound images) and an RNN for textual feature extraction. The CNN extracts image features, while the RNN encodes text features, merging them to generate captions for anatomical structures. The model is evaluated with BLEU and ROUGE-L metrics and produces relevant and descriptive captions for educating sonography trainees and patients.\\
In \cite{Zeng2020}, a new method for ultrasound image captioning based on region detection is introduced to improve disease content analysis. The model detects and encodes focus areas in ultrasound images and then uses LSTM to generate descriptive text. This method increases accuracy in focus area detection and achieves higher BLEU-1 and BLEU-2 scores with fewer parameters and faster runtimes than traditional models. \\
Expanding on incorporating additional data types, \cite{Zeng2020b} introduces a Semantic Fusion Network to improve the accuracy of medical image diagnostic reports by integrating pathological information. This network comprises a lesion area detection model that extracts visual and pathological data and a diagnostic generation model that combines this information to produce reports. This method enhances the accuracy of generated reports, showing a 1.2\% increase in the ultrasound image dataset compared to models relying solely on visual features.\\
In a similar vein of enhancing multimodal integration, \cite{Yang2021} introduces an Adaptive Multimodal Attention network to generate high-quality medical image reports. The model employs a multilabel classification network to predict local properties of ultrasound images, using their word embeddings as semantic features. It integrates semantic and adaptive attention mechanisms with a sentinel gate to balance focus between visual features and language model memories. This approach enhances report accuracy and robustness, outperforming baseline models in capturing key local properties.\\
Addressing the challenge of small datasets, \cite{Alsharid2022} presents a weakly-supervised method to enhance image captioning models using a large anatomically-labeled image classification dataset. This encoder-decoder model generates pseudo-captions for unlabeled images, creating an augmented dataset that significantly improves fetal ultrasound image captioning. This approach nearly doubles BLEU-1 and ROUGE-L scores, saving time on manual annotations and improving model performance in communicating information to laypersons.\\
In \cite{Deng2024}, a transformer-based model is proposed to generate descriptive ultrasound images of lymphoma, providing auxiliary guidance for sonographers. The model integrates deep stable learning to eliminate feature dependencies and includes a memory module for enhanced semantic modeling. Using a nonlinear feature decorrelation method, this approach visualizes cross-attention for interpretability and focuses on lymphoma features over the background. The result is a more accurate and detailed depiction of lymphoma in ultrasound images.\\
To further improve automatic report generation, \cite{Li2024} introduces a framework utilizing both unsupervised and supervised learning to align visual and textual features. Unsupervised learning extracts knowledge from text reports, guiding the model, while a global semantic comparison mechanism ensures accurate, comprehensive reports. Tested on three large datasets (breast, thyroid, liver), the method outperforms other approaches without needing manual disease labels, enhancing efficiency and accessibility.
\section{Methodology and Proposed Approach}\label{sec3}
This study presents a novel uterine ultrasound image captioning system. To achieve this, we first gathered a diverse dataset of uterine ultrasound images and meticulously annotated them with precise medical terminology, covering women of various ages and pregnancy stages. Our approach involved rigorous data preprocessing for both images and text, followed by feature extraction using pre-trained CNN-based models. Finally, we implemented our proposed deep learning model, CNN-BiLGRU. Detailed descriptions of each module follow in the subsequent sections.
\subsection{Data Collection and Annotation}
Our dataset focuses specifically on gynecology, the branch of medicine that deals with women’s health. We created a dataset that delves deeper into the details of gynecological imaging to address the specific challenges doctors face when diagnosing gynecological problems. This section details collecting and annotating medical images to train the medical image captioning model.

Our research utilized a dataset of ultrasound images exceeding 500 in number (505 images). Data collection involved acquiring ultrasound images from three main sources (see Figure \ref{fig1}). Internally, we gathered 214 images obtained directly from the Sonoscape SS1-8000 machine. Each image has a dimension of 1024x768 pixels (width: 1024 pixels, height: 768 pixels) and is stored in JPG format.
\begin{figure}[h!]
	\centering
	\includegraphics[width=0.8\columnwidth]{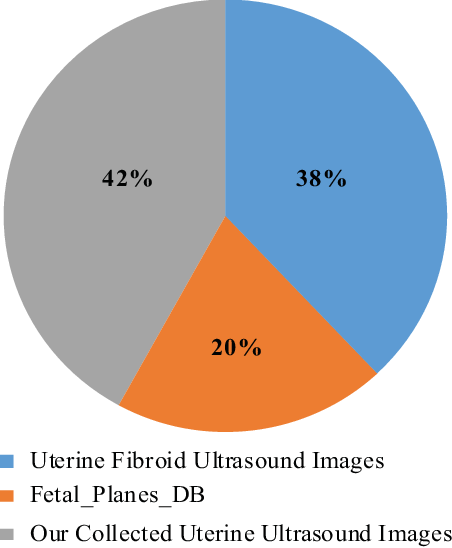}
	\caption{Proportion of used uterine ultrasound images by source.}
	\label{fig1}
\end{figure}
Externally, we incorporated data from publicly available datasets to enrich this collection and capture a wider range of variations. From the Mendeley repository, which offered a rich collection of nearly 1,500 fetal ultrasound images (uterine fibroid ultrasound images \cite{mendeley}), we collaborated with experts to meticulously review and select a subset of 191 images that best aligned with our research goals. This selection process involved eliminating images with repetitive features, poor capture of the region of interest, or other factors that could negatively impact model training. Additionally, the Zenodo \cite{zenodo} dataset (Fetal Planes DB) provided 450 images, meticulously organized to include four images per patient, each representing the standard fetal planes of the abdomen, brain, femur, and thorax. From this collection, we selected a subset of 100 images that best aligned with our research goals, ensuring comprehensive coverage of fetal anatomy across multiple datasets.\\
In the form of captions, annotations were then added to each image in our dataset. These captions captured key features and findings within the ultrasound images, including identifying anatomical structures such as the stomach, umbilical vein, femur bones, and brain ventricles and noting potential abnormalities such as dilated organs or fluid pockets in the brain. We communicated closely with experts during the annotation process to ensure accuracy and quality. This collaboration helped us resolve image-related issues and made our dataset more valuable for analysis and research.

\subsection{Data Pre-processing}
Data preprocessing is crucial for ensuring the quality and usability of data for subsequent analysis and modeling \cite{Boulesnane2022}. Our study encompasses rigorous processing of images and text to enhance data integrity and relevance.

Image processing plays a crucial role in refining collected data. Initially, we analyze and filter the images to align with project requirements. The first step involves cropping the images to focus on the Region of Interest (ROI). Upon reading each ultrasound image, we convert it to grayscale if it is in color. Subsequently, we determine the cropping points by identifying significant changes in pixel intensity from the image center toward its edges. This process begins by calculating the mean intensity column-wise for both the right and left halves of the image, as depicted in Figure \ref{fig:sub1}. Peaks in these intensity profiles highlight areas of interest, and points where intensity drops below a predefined threshold (5\% of peak value) denote edges of the ROI (see Figure \ref{fig:sub2}). Using these change points, we derive precise cropping coordinates to isolate the ROI (Figure \ref{fig:sub3}).
\begin{figure*}[h!]
	\centering
	\begin{subfigure}[b]{\textwidth}
		\centering
		\includegraphics[width=0.9\textwidth,height=7cm]{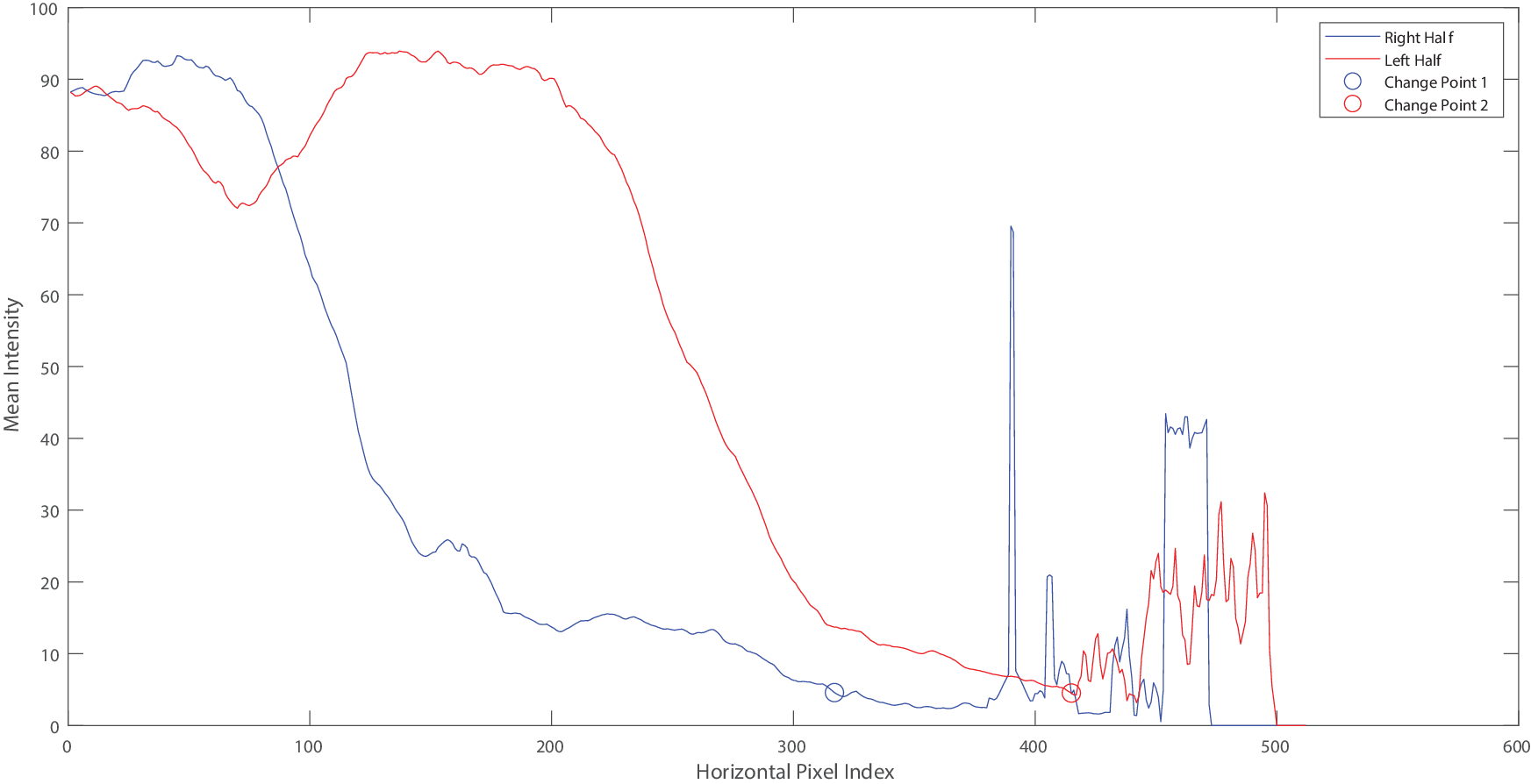}
		\caption{Mean intensity variation across image width.}
		\label{fig:sub1}
	\end{subfigure}
	\bigskip
	
	\begin{subfigure}[b]{0.45\textwidth}
		\centering
		\includegraphics[width=\textwidth,height=5cm]{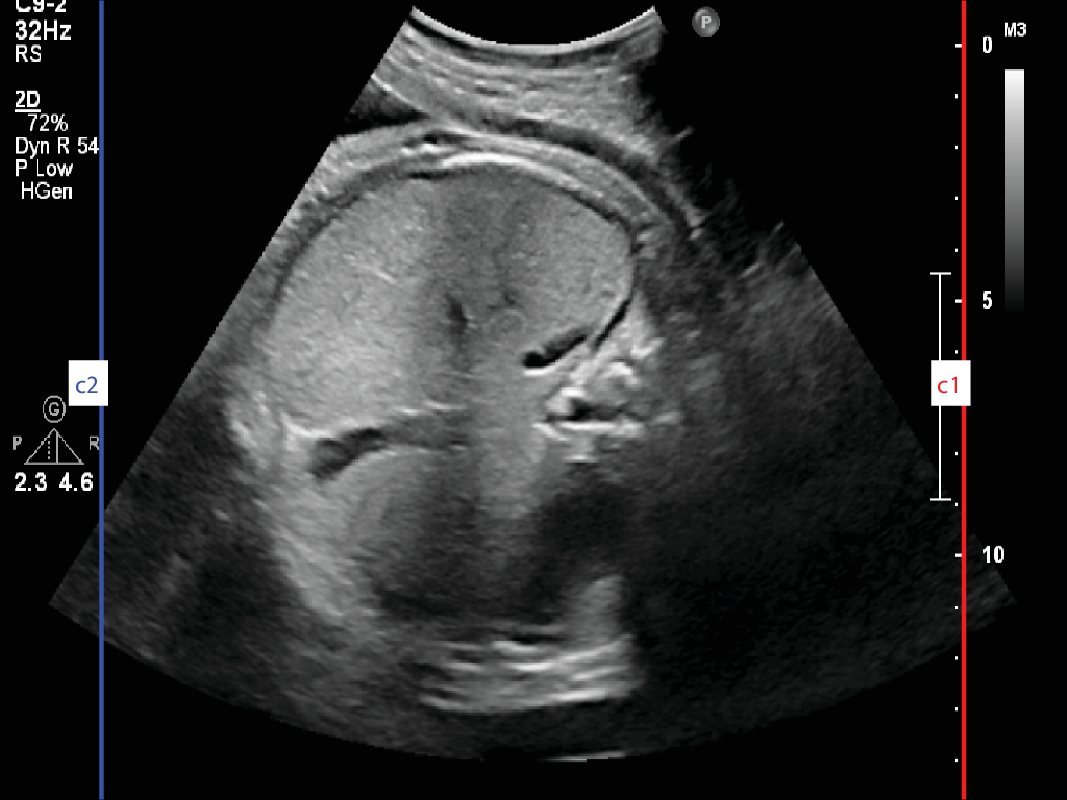}
		\caption{Original image with crop lines.}
		\label{fig:sub2}
	\end{subfigure}
	\hspace{1em}
	\begin{subfigure}[b]{0.45\textwidth}
		\centering
		\includegraphics[width=\textwidth,height=5cm]{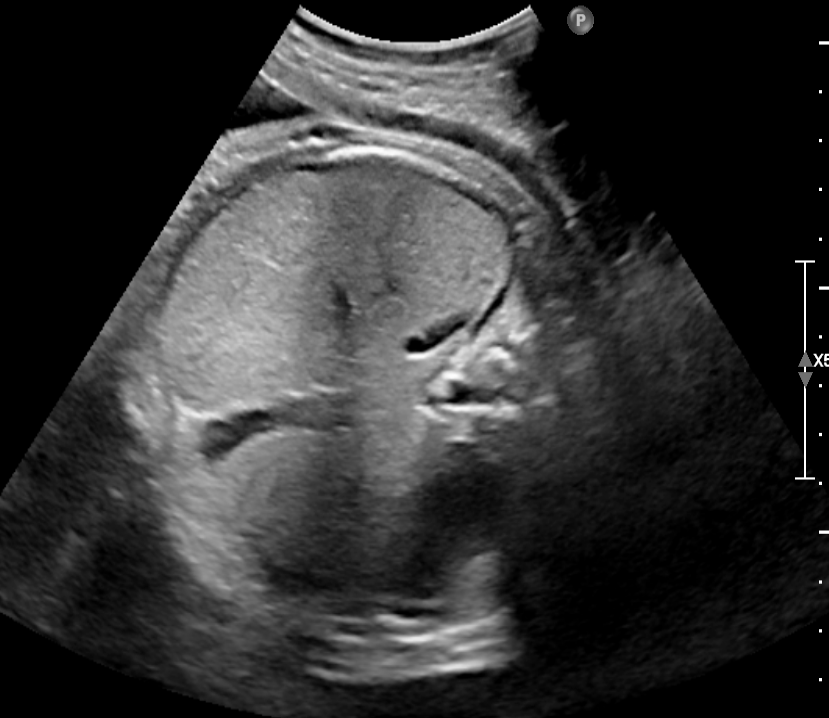}
		\caption{ROI-cropped image.}
		\label{fig:sub3}
	\end{subfigure}
	\caption{Original image to the region of interest.}
	\label{fig:overall}
\end{figure*}
Post-cropping, all images are resized uniformly to 224x224 pixels, a standard size compatible with many pre-trained neural networks. Each resized image instance is then converted into a Numpy array and normalized. Normalization involves scaling pixel values from 0 to 255 to a normalized range of 0 to 1 by dividing each pixel value by 255.\\

After completing the image processing and preparing the images, we focused on processing the text captions associated with each image. These captions were derived from expert-provided medical descriptions and were systematically linked to their corresponding image file names within an Excel file. To enhance the text data for subsequent analysis, we applied NLP techniques \cite{Boulesnane2022b}:
\begin{itemize}
	\item Convert to Lowercase: All sentences were converted to lowercase to maintain consistency and reduce variability across the dataset.
	\item Remove Punctuation: Punctuation marks were systematically removed to simplify the text and emphasize the words.
	\item  Remove Single Letters: Single letters such as 'l', 's', 'a', and 'à' were removed, as they typically do not contribute significant meaning in medical contexts.
	\item  Remove Extra Spaces: Any extraneous spaces within the text were eliminated to ensure uniform spacing and improve text clarity.
	\item Add Start and End Tags: Special tags \textless{}START\textgreater{} and \textless{}END\textgreater{}
	were appended to the beginning and end of each sentence. These tags serve as markers during subsequent text processing and modeling to delineate sentence boundaries effectively.
\end{itemize}
\subsection{Feature Extraction}
Feature extraction is crucial in identifying and describing pertinent information within patterns \cite{Salau2019}. This process facilitates pattern classification by establishing a structured and systematic approach. This phase focuses on deriving meaningful numerical representations from text descriptions and ultrasound images.

Text feature extraction involves converting textual data, such as medical reports and captions, into a format suitable for machine learning models. Initially, we employ a tokenizer to create a dictionary of word indices from our text data. This step allows us to determine the vocabulary size, represent the total number of unique words in the dataset, and identify the longest caption's length. Subsequently, we construct a vocabulary of unique words to map each word to its corresponding index. Shorter sequences are padded with zeros to ensure uniform input sequence lengths (captions), as neural networks require consistent input dimensions.
\begin{figure*}[b!]
	\centering
	\includegraphics[width=0.9\textwidth]{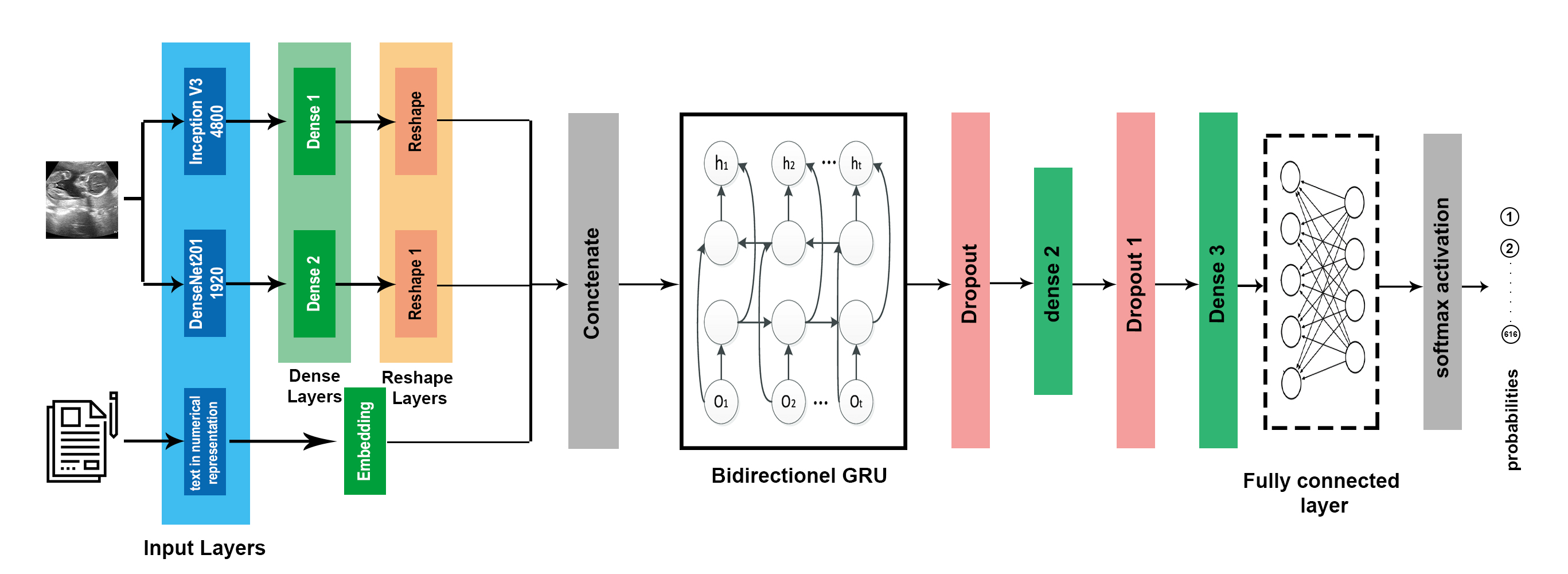}
	\caption{Architecture of the proposed CNN-BiGRU model.}
	\label{fig3}
\end{figure*}

In addition to text, image feature extraction in this study utilizes advanced convolutional neural network architectures, namely Inception V3 and DenseNet201. These models have been pre-trained on the vast ImageNet dataset \cite{Krizhevsky2017}, which consists of millions of annotated images across thousands of categories. The key advantage of using these pre-trained models is their ability to capture intricate patterns and hierarchical representations within images.
\subsection{Proposed Uterine Ultrasound Image Captioning Model}
The proposed uterine ultrasound image captioning model aims to generate meaningful and accurate captions for medical ultrasound images of the uterus. To achieve high accuracy, the system architecture incorporates several advanced components. Our dataset consists of 505 images specifically selected to represent various uterine ultrasound scans commonly encountered in clinical practice.

As shown in Figure \ref{fig3}, the model's architecture begins with three input layers. The first input layer receives features extracted from a DenseNet201 model, shaped as (None, 1920). The second input layer obtains features from an InceptionV3 model, shaped as (None, 4800). The third input layer receives tokenized text sequences with a (None, 54) shape, where 54 represents the maximum caption length. The image features from the DenseNet201 and InceptionV3 models pass through dense layers that reduce their dimensionality to (None, 256). These outputs are then reshaped into (None, 1, 256) using reshape layers. Meanwhile, the tokenized text sequences are embedded, resulting in fixed-size tensor vectors (None, 54, 256).\\
The embeddings are concatenated with the reshaped image features, and the combined data is fed into a bidirectional GRU layer, which processes sequential data bidirectionally and produces an output shape of (None, 256). A dropout layer with a dropout rate of 0.5 is applied to prevent overfitting. The output is then passed through an intermediate dense layer that reduces the dimensionality to (None, 128), followed by another dropout layer with the same rate.\\
Finally, a dense layer with a softmax activation function generates the final output, which has a shape of (None, 626). This represents the predicted caption probabilities for each word in the vocabulary. By integrating image and text features, this architecture produces accurate and informative captions for uterine ultrasound images, thereby enhancing medical diagnosis and treatment planning.
\section{Experiments}\label{sec4}
In this section, we detail the experimental setup and analyze the results of our image captioning model. Regarding configuration, we divided the dataset, allocating 85\% for training and 15\% for testing (validation). Furthermore, the model parameters were configured with the Adam optimizer, a batch size of 16, and an early stopping patience of 10 epochs.\\
We analyze the performance of the proposed model and compare it with other baseline models to ensure a comprehensive evaluation. For this, we employed several metrics to evaluate the generated captions, including BLEU \cite{papineni2002bleu} and ROUGE \cite{lin2004rouge} scores. BLEU scores (BLEU1, BLEU2, BLEU3, BLEU4) are widely used in machine translation tasks, and they measure the similarity between the generated captions and the ground truth references using n-grams. ROUGE scores (ROUGE1, ROUGE2, ROUGEL) are standard in text summarization. ROUGE1 and ROUGE2 measure the recall of unigrams and bigrams, respectively, while ROUGEL evaluates the recall of the longest common subsequences between the generated and reference captions. These metrics comprehensively assess the model's performance in generating accurate and relevant captions for uterine ultrasound images.
\subsection{Performance Analysis of the Proposed CNN-BiGRU Model}
This analysis evaluates the performance of our CNN-BiGRU model, which leverages the powerful feature extraction capabilities of pre-trained Inception V3 and DenseNet201 architectures, combined with the temporal sequence modeling strength of BiGRU, for captioning uterine ultrasound images.
\begin{figure*}[h!]
	\centering
	\includegraphics[width=\textwidth]{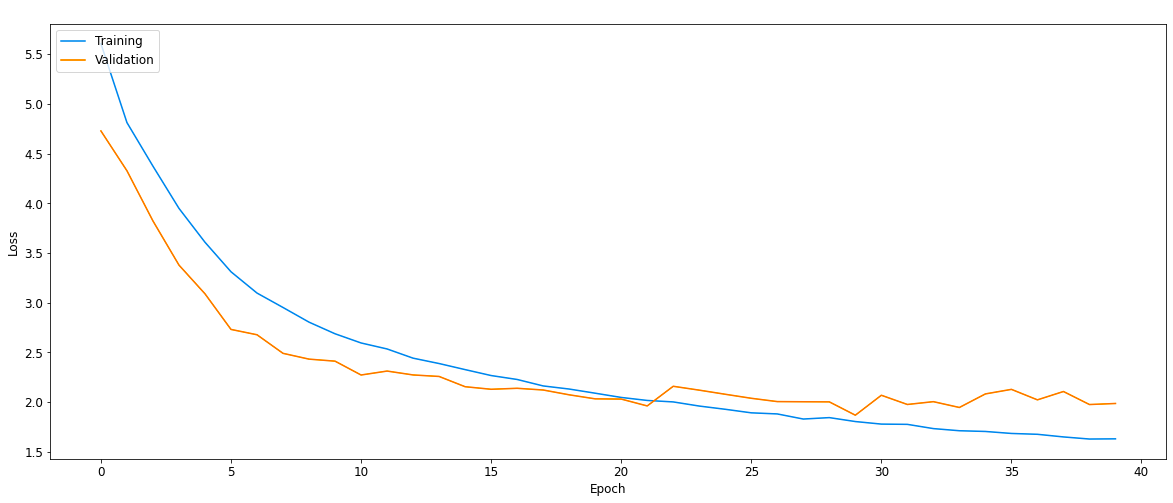}
	\caption{Loss curve of our proposed CNN-BiGRU model.}
	\label{fig4}
\end{figure*}

Figure \ref{fig4} displays the learning curves for the loss of the proposed CNN-BiGRU model during both the training and validation phases. These curves indicate that the model was trained appropriately, with no signs of overfitting. The training and validation losses were closely aligned throughout the training process, a positive indicator of the model's generalization capability.\\
Specifically, at epoch 29, the model achieved a training loss of 1.64 and a validation loss of 1.86. These loss values suggest that the model effectively learned the underlying patterns in the data while maintaining a balance between fitting the training data and generalizing it to unseen validation data. The training was terminated at epoch 39 due to the early stopping criterion, set with a patience of 10 epochs. This means that the model stopped training when there was no significant improvement in the validation loss for 10 consecutive epochs, thereby preventing overfitting and ensuring that the model maintained its performance on the validation set.\\
Achieving a low loss in image captioning is challenging because it requires understanding visual content, recognizing objects and relationships, and translating this into coherent text. Variability in descriptions and sequential dependency in caption generation add complexity. Additionally, aligning visual features with textual representations involves bridging the gap between two different data modalities (i.e., images and text).
\subsection{Comparison with Baseline Models}
Our study explored various architectures integrated with different processing layers to generate captions for uterine ultrasound images. We primarily focused on using DenseNet201 and InceptionV3 models for feature extraction, followed by BiGRU, as well as baseline models such as Unidirectional GRU (UniGRU), Bidirectional Long Short-Term Memory (BiLSTM), and Unidirectional Long Short-Term Memory (UniLSTM) networks.
\begin{figure*}[b!]
	\centering
	\begin{subfigure}[b]{0.48\textwidth}
		\centering
		\includegraphics[width=\textwidth,height=5cm]{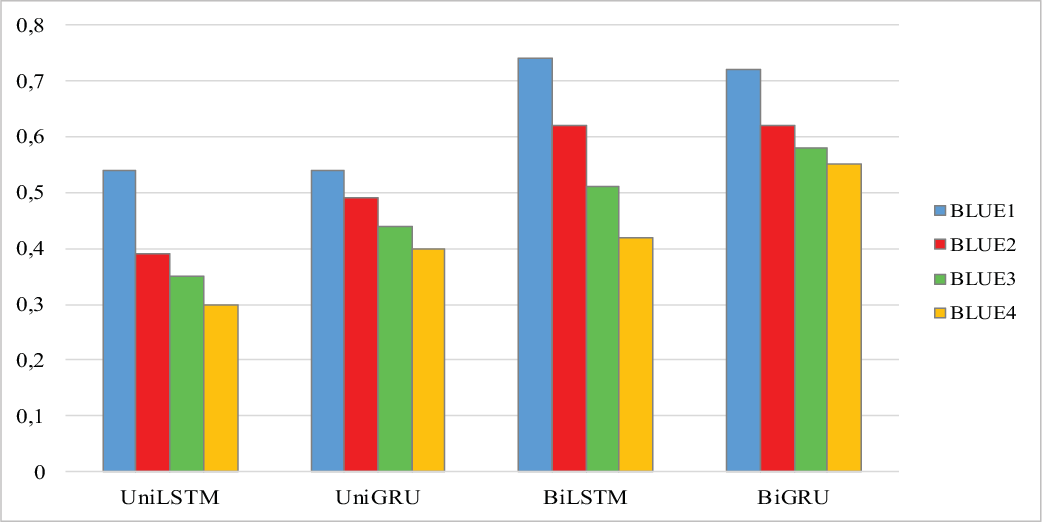}
		\caption{}
		\label{fig5-1}
	\end{subfigure}
	\hfill
	\begin{subfigure}[b]{0.48\textwidth}
		\centering
		\includegraphics[width=\textwidth,height=5cm]{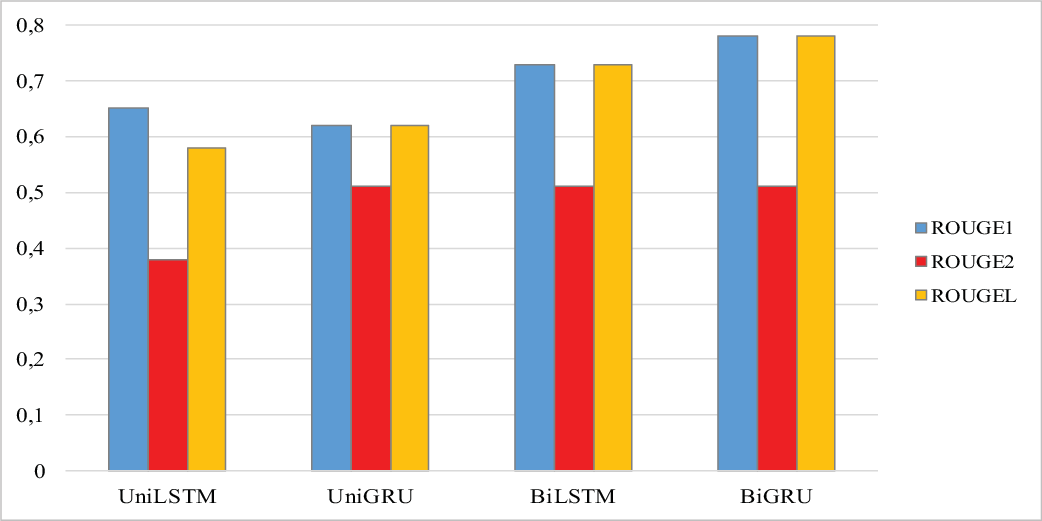}
		\caption{}
		\label{fig5-2}
	\end{subfigure}
	\bigskip
	\begin{subfigure}[b]{0.6\textwidth}
		\centering
		\includegraphics[width=\textwidth,height=5cm]{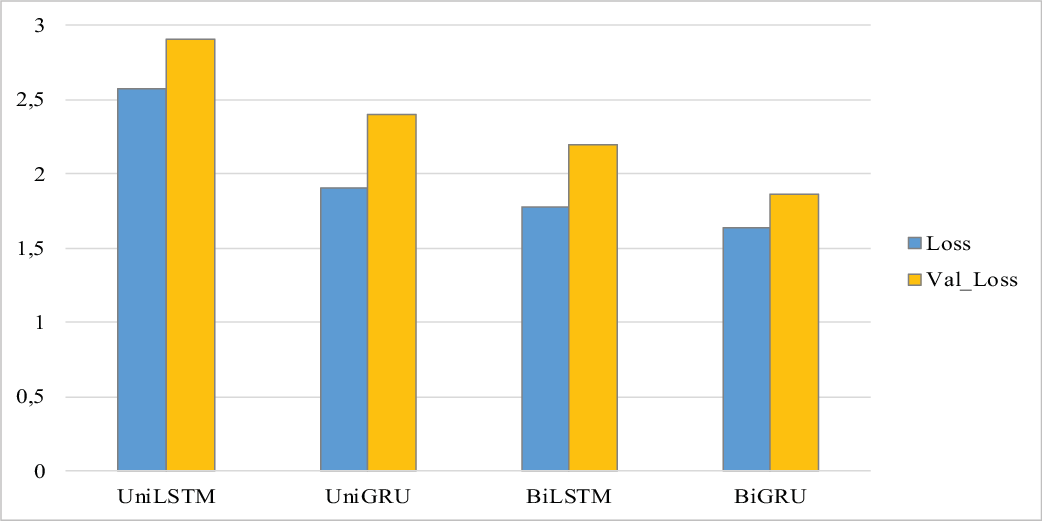}
		\caption{}
		\label{fig5-3}
	\end{subfigure}
	
	\caption{Comparative performance analysis of the proposed model and the baseline models. (a) Comparison of BLEU scores. (b) Comparison of ROUGE scores. (c) Comparison of training and validation loss.}
	\label{fig5}
\end{figure*}

To provide a comprehensive comparison, we evaluated the performance of these models using several metrics, including BLEU and ROUGE scores. Higher scores indicate better performance. As depicted in Figure \ref{fig5-1}, the BLEU scores for our models showed that BiGRU and BiLSTM outperformed the baseline models UniGRU and UniLSTM, with BiGRU achieving the highest BLEU-4 score of 0.55. At the same time, the ROUGE scores highlighted BiGRU as the best performer, with a ROUGE-L score of 0.78, as shown in Figure \ref{fig5-2}.\\
We also analyzed the training loss and validation loss for the selected models (see Figure \ref{fig5-3}). The values of "Loss", which refers to the training loss calculated on the training dataset, and "Val\_Loss", which stands for validation loss calculated on the dataset, are important indicators of model performance. Our results showed that the BiGRU model had the lowest loss values (as shown before in Figure \ref{fig4}), with a training loss of 1.64 and a validation loss of 1.86, indicating its robustness and effectiveness in generating accurate and contextually appropriate captions for uterine ultrasound images.
\begin{table*}[b!]
	\centering
	\caption{Sample outputs comparing reference captions with captions generated by the proposed model.}
	\label{tab:comparison}
	\begin{tabularx}{\textwidth}{|c|c|c|}
		\hline
		\textbf{Uterine Ultrasound Image} & \textbf{Reference Caption} & \textbf{Generated Caption} \\
		\hline
		\begin{minipage}[c][4cm][c]{0.26\textwidth}
			\centering
			\includegraphics[width=\linewidth]{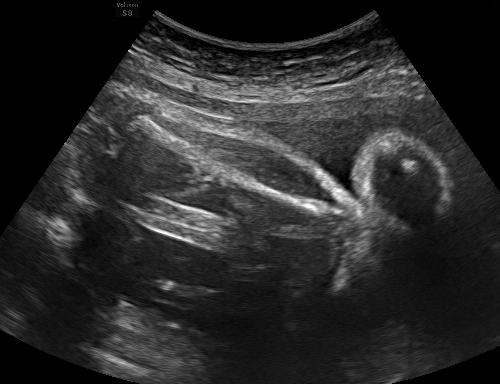}
		\end{minipage} & 
		\begin{minipage}[c][4cm][c]{0.3311\textwidth}
			a white straight line at the top center that represents the femur bone it is possible to calculate the femur length the knee is straight
		\end{minipage} & 
		\begin{minipage}[c][4cm][c]{0.3311\textwidth}
			a white line at the top center that represents the femur bone it is possible to calculate the femur length the knee is straight
		\end{minipage} \\
		\hline
		\begin{minipage}[c][4cm][c]{0.26\textwidth}
			\centering
			\includegraphics[width=\linewidth]{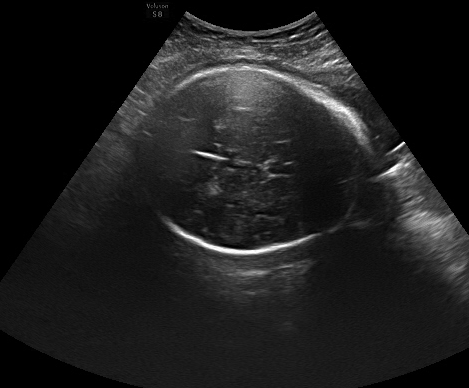}
		\end{minipage} & 
		\begin{minipage}[c][4cm][c]{0.3333\textwidth}
			a large slightly oval circle that represents the cranial contour of the fetus inside it is possible to see the cavum of the septum pellucidum on the right only but it is possible to calculate the biparietal diameter
		\end{minipage} & 
		\begin{minipage}[c][4cm][c]{0.3333\textwidth}
			a large slightly oval circle that represents the cranial contour of the fetus the cavum of the septum pellucidum can be seen on the right it is possible to calculate the biparietal diameter
		\end{minipage} \\
		\hline
	\end{tabularx}
\end{table*}

BiGRU's superiority can be attributed to its ability to capture bidirectional dependencies in sequences, which is crucial for understanding and generating coherent and contextually accurate captions. Unlike the baseline unidirectional models, BiGRU can process the sequence data in both forward and backward directions, leading to a more comprehensive understanding of the temporal context. This capability is especially beneficial for complex tasks like image captioning, where the relationship between different parts of an image and the corresponding text must be accurately captured.

Consequently, the CNN-BiGRU model outperformed the other baseline models regarding BLEU and ROUGE scores and demonstrated lower loss values, underscoring its effectiveness in this application. Moreover, Table \ref{tab:comparison} provides further evidence by illustrating sample outputs that compare reference captions with captions generated by the CNN-BiGRU model. This comparison vividly showcases the model's ability to generate high-quality captions through its robust bidirectional processing capabilities.
\section{Conclusion and Future Work}\label{sec5}
In this study, we successfully developed a deep learning-based medical image captioning system designed explicitly for uterine ultrasound images, utilizing a CNN-BiGRU architecture. Our model effectively combined the image feature extraction capabilities of pre-trained CNNs (InceptionV3 and DenseNet201) with the sequential processing strength of a Bidirectional Gated Recurrent Unit network. This hybrid approach demonstrated superior performance to baseline models, achieving higher BLEU and ROUGE scores and maintaining low training and validation losses. The generated captions were accurate and informative, improving the interpretability of complex uterine ultrasound images.\\
The findings from our research underscore the potential of deep learning techniques in enhancing diagnostic precision and efficiency in obstetrics and gynecology. By automating the captioning process, our model assists medical professionals in making timely and accurate diagnoses, which can lead to improved patient outcomes. The effectiveness of our approach not only contributes to the field of medical image captioning but also highlights the broader applicability of AI in healthcare.

Building upon the promising results of our CNN-BiGRU model, several avenues for future research and development are suggested. Expanding and diversifying the dataset with more uterine ultrasound images from diverse sources will enhance the model's robustness and generalizability. Exploring advanced architectures and optimization techniques, such as attention mechanisms or transformer-based models, may improve caption quality. Developing a real-time image captioning system for clinical integration, creating user interfaces with feedback mechanisms, and integrating multimodal medical data can provide more comprehensive diagnostic tools. 

\section*{Data Availability}
The data supporting the findings of this study can be obtained from the corresponding authors upon reasonable request.

\bibliographystyle{IEEEtran}
\bibliography{IEEEabrv,ref}

\begin{thebibliography}{10}
\providecommand{\url}[1]{#1}
\csname url@samestyle\endcsname
\providecommand{\newblock}{\relax}
\providecommand{\bibinfo}[2]{#2}
\providecommand{\BIBentrySTDinterwordspacing}{\spaceskip=0pt\relax}
\providecommand{\BIBentryALTinterwordstretchfactor}{4}
\providecommand{\BIBentryALTinterwordspacing}{\spaceskip=\fontdimen2\font plus
\BIBentryALTinterwordstretchfactor\fontdimen3\font minus
  \fontdimen4\font\relax}
\providecommand{\BIBforeignlanguage}[2]{{%
\expandafter\ifx\csname l@#1\endcsname\relax
\typeout{** WARNING: IEEEtran.bst: No hyphenation pattern has been}%
\typeout{** loaded for the language `#1'. Using the pattern for}%
\typeout{** the default language instead.}%
\else
\language=\csname l@#1\endcsname
\fi
#2}}
\providecommand{\BIBdecl}{\relax}
\BIBdecl

\bibitem{Haidekker2013}
M.~A. Haidekker, \emph{Medical Imaging Technology}.\hskip 1em plus 0.5em minus
  0.4em\relax Springer New York, 2013.

\bibitem{bradley2008history}
W.~G. Bradley, ``History of medical imaging,'' \emph{Proceedings of the
  American Philosophical Society}, vol. 152, no.~3, pp. 349--361, 2008.

\bibitem{Obuchowicz2024}
R.~Obuchowicz, M.~Strzelecki, and A.~Pi{\'o}rkowski, \emph{Artificial
  Intelligence in Medical Imaging and Image Processing}.\hskip 1em plus 0.5em
  minus 0.4em\relax MDPI, May 2024.

\bibitem{Suzuki2017}
K.~Suzuki, ``Overview of deep learning in medical imaging,'' \emph{Radiological
  Physics and Technology}, vol.~10, no.~3, p. 257–273, Jul. 2017.

\bibitem{Kaul2021}
D.~Kaul, H.~Raju, and B.~K. Tripathy, \emph{Deep Learning in Healthcare}.\hskip
  1em plus 0.5em minus 0.4em\relax Springer International Publishing, Aug.
  2021, p. 97–115.

\bibitem{Beddiar2022}
D.-R. Beddiar, M.~Oussalah, and T.~Sepp\"{a}nen, ``Automatic captioning for
  medical imaging (mic): a rapid review of literature,'' \emph{Artificial
  Intelligence Review}, vol.~56, no.~5, p. 4019–4076, Sep. 2022.

\bibitem{Xu2023}
L.~Xu, Q.~Tang, J.~Lv, B.~Zheng, X.~Zeng, and W.~Li, ``Deep image captioning: A
  review of methods, trends and future challenges,'' \emph{Neurocomputing},
  vol. 546, p. 126287, Aug. 2023.

\bibitem{LevienaiseObadia1999}
B.~Levienaise-Obadia and A.~Gee, ``Adaptive segmentation of ultrasound
  images,'' \emph{Image and Vision Computing}, vol.~17, no.~8, p. 583–588,
  Jun. 1999.

\bibitem{Chen2016}
H.~Chen, Y.~Zheng, J.-H. Park, P.-A. Heng, and S.~K. Zhou, \emph{Iterative
  Multi-domain Regularized Deep Learning for Anatomical Structure Detection and
  Segmentation from Ultrasound Images}.\hskip 1em plus 0.5em minus 0.4em\relax
  Springer International Publishing, 2016, p. 487–495.

\bibitem{Zeng2018}
X.-H. Zeng, B.-G. Liu, and M.~Zhou, ``Understanding and generating ultrasound
  image description,'' \emph{Journal of Computer Science and Technology},
  vol.~33, no.~5, p. 1086–1100, Sep. 2018.

\bibitem{Wang2024}
\BIBentryALTinterwordspacing
X.~Wang, G.~Figueredo, R.~Li, W.~E. Zhang, W.~Chen, and X.~Chen, ``A survey of
  deep learning-based radiology report generation using multimodal data,''
  2024. [Online]. Available: \url{https://arxiv.org/abs/2405.12833}
\BIBentrySTDinterwordspacing

\bibitem{Alsharid2019}
M.~Alsharid, H.~Sharma, L.~Drukker, P.~Chatelain, A.~T. Papageorghiou, and
  J.~A. Noble, \emph{Captioning Ultrasound Images Automatically}.\hskip 1em
  plus 0.5em minus 0.4em\relax Springer International Publishing, 2019, p.
  338–346.

\bibitem{Zeng2020}
X.~Zeng, L.~Wen, B.~Liu, and X.~Qi, ``Deep learning for ultrasound image
  caption generation based on object detection,'' \emph{Neurocomputing}, vol.
  392, p. 132–141, Jun. 2020.

\bibitem{Zeng2020b}
X.~Zeng, L.~Wen, Y.~Xu, and C.~Ji, ``Generating diagnostic report for medical
  image by high-middle-level visual information incorporation on double deep
  learning models,'' \emph{Computer Methods and Programs in Biomedicine}, vol.
  197, p. 105700, Dec. 2020.

\bibitem{Yang2021}
S.~Yang, J.~Niu, J.~Wu, Y.~Wang, X.~Liu, and Q.~Li, ``Automatic ultrasound
  image report generation with adaptive multimodal attention mechanism,''
  \emph{Neurocomputing}, vol. 427, p. 40–49, Feb. 2021.

\bibitem{Alsharid2022}
M.~Alsharid, H.~Sharma, L.~Drukker, A.~T. Papageorgiou, and J.~A. Noble,
  \emph{Weakly Supervised Captioning of Ultrasound Images}.\hskip 1em plus
  0.5em minus 0.4em\relax Springer International Publishing, 2022, p.
  187–198.

\bibitem{Deng2024}
J.~Deng, D.~Chen, C.~Zhang, and Y.~Dong, ``Generating lymphoma ultrasound image
  description with transformer model,'' \emph{Computers in Biology and
  Medicine}, vol. 174, p. 108409, May 2024.

\bibitem{Li2024}
J.~Li, T.~Su, B.~Zhao, F.~Lv, Q.~Wang, N.~Navab, Y.~Hu, and Z.~Jiang,
  ``Ultrasound report generation with cross-modality feature alignment via
  unsupervised guidance,'' 2024.

\bibitem{mendeley}
T.~Yang, ``Uterine fibroid ultrasound images,'' 2023.

\bibitem{zenodo}
X.~P. Burgos-Artizzu, D.~Coronado-Gutierrez, B.~Valenzuela-Alcaraz,
  E.~Bonet-Carne, E.~Eixarch, F.~Crispi, and E.~Gratacós,
  ``{FETAL\_PLANES\_DB: Common maternal-fetal ultrasound images},'' Jun. 2020.

\bibitem{Boulesnane2022}
A.~Boulesnane, S.~Meshoul, and K.~Aouissi, ``Influenza-like illness detection
  from arabic facebook posts based on sentiment analysis and 1d convolutional
  neural network,'' \emph{Mathematics}, vol.~10, no.~21, p. 4089, Nov. 2022.

\bibitem{Boulesnane2022b}
A.~Boulesnane, Y.~Saidi, O.~Kamel, M.~M. Bouhamed, and R.~Mennour, ``Dzchatbot:
  A medical assistant chatbot in the algerian arabic dialect using seq2seq
  model,'' in \emph{2022 4th International Conference on Pattern Analysis and
  Intelligent Systems (PAIS)}.\hskip 1em plus 0.5em minus 0.4em\relax IEEE,
  Oct. 2022.

\bibitem{Salau2019}
A.~O. Salau and S.~Jain, ``Feature extraction: A survey of the types,
  techniques, applications,'' in \emph{2019 International Conference on Signal
  Processing and Communication (ICSC)}.\hskip 1em plus 0.5em minus 0.4em\relax
  IEEE, Mar. 2019.

\bibitem{Krizhevsky2017}
A.~Krizhevsky, I.~Sutskever, and G.~E. Hinton, ``Imagenet classification with
  deep convolutional neural networks,'' \emph{Communications of the ACM},
  vol.~60, no.~6, p. 84–90, May 2017.

\bibitem{papineni2002bleu}
K.~Papineni, S.~Roukos, T.~Ward, and W.-J. Zhu, ``Bleu: a method for automatic
  evaluation of machine translation,'' in \emph{Proceedings of the 40th annual
  meeting of the Association for Computational Linguistics}, 2002, pp.
  311--318.

\bibitem{lin2004rouge}
C.-Y. Lin, ``Rouge: A package for automatic evaluation of summaries,'' in
  \emph{Text summarization branches out}, 2004, pp. 74--81.

\end{thebibliography}

\begin{IEEEbiography}[{\includegraphics[width=1in,height=1.25in,clip,keepaspectratio]{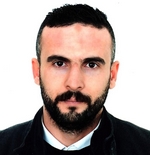}}]{Abdennour Boulesnane}
 		\noindent  received his B.S. degree from Mentouri Constantine University (UMC) and M.S. degree from the Department of Computer Science and Application, Constantine 2 University, Constantine, Algeria, respectively, in 2011 and 2013. He received his PhD degree in Computer Science from Abdelhamid Mehri Constantine 2 University, Algeria, in 2017. Currently, he holds the position of Associate Professor of Computer Science within the Department of Medicine at the University of Salah Boubnider - Constantine 3. Additionally, he is a founding member of the BIOSTIM Research Laboratory at the same academic institution. His research interests include computational medicine, optimization and metaheuristics, Computer vision, NLP, Machine Learning, and Data Science.
\end{IEEEbiography}
 \vspace{-8cm}
\begin{IEEEbiography}[{\includegraphics[width=1in,height=1.25in,clip,keepaspectratio]{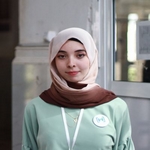}}]{Boutheina Mokhtari}
 \noindent  received her B.S. and M.S. degrees from the Department of IFA, Faculty of NTIC, Abdelhamid Mehri University, Constantine, Algeria, respectively, in 2022 and 2024. Her areas of interest are Artificial Intelligence, Deep Learning, and Computer Vision.
\end{IEEEbiography}	
 \vspace{-8cm}
\begin{IEEEbiography}[{\includegraphics[width=1in,height=1.25in,clip,keepaspectratio]{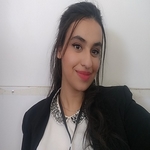}}]{Oumnia Rana Segueni}
\noindent earned her Bachelor's and Master's degrees from the Department of IFA, Faculty of NTIC, at Abdelhamid Mehri University in Constantine, Algeria, in 2022 and 2024, respectively. Her research interests include Artificial Intelligence, Deep Learning, and Computer Vision.
 \end{IEEEbiography}
 \vspace{-8cm}
\begin{IEEEbiography}[{\includegraphics[width=1in,height=1.25in,clip,keepaspectratio]{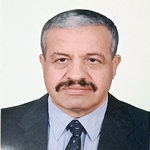}}]{Slimane Segueni} 
\noindent completed his residency in gynecology and obstetrics in 1988 in Constantine, Algeria. Following his residency, he served as an assistant professor at SMK Maternity Hospital. He then moved to Ghardaia, where he worked as the chief physician for two years. In 1991, he opened his private practice in Chelghoum, where he has been serving the community ever since. Notably, he volunteered his medical expertise during the Chlef earthquakes, demonstrating his commitment to public service even before completing his residency.
 	\end{IEEEbiography}
\end{document}